\documentclass[letterpaper, 10 pt, conference]{ieeeconf}
\pdfoutput=1

\IEEEoverridecommandlockouts

\usepackage[style=ieee, doi=false, url=false, eprint=false, isbn=false, natbib=true, mincitenames=1, maxcitenames=1, maxbibnames=99]{biblatex}
\usepackage{hyperref}
\usepackage[subpreambles=true]{standalone}
\usepackage{siunitx}
\usepackage{caption}
\usepackage{graphicx}
\usepackage{subcaption}
\usepackage{amsmath}
\usepackage{amssymb}
\usepackage{mathrsfs}
\usepackage{booktabs}
\usepackage{flushend}

\graphicspath{{../}{./}{./figures/}}
\usepackage[capitalize]{cleveref}

\newcommand{\Initialize}{\textit{Initialize}}
\newcommand{\Step}{\textit{Step}}
\newcommand{\IsTerminal}{\textit{IsTerminal}}
\newcommand{\IsFailure}{\textit{IsFailure}}

\title{\LARGE \bf
How Do We Fail? Stress Testing Perception in Autonomous Vehicles
}

\author{Harrison Delecki$^{1}$, Masha Itkina$^{1}$,  Bernard Lange$^{1}$,  Ransalu Senanayake$^{2}$,  Mykel J. Kochenderfer$^{1}$
\thanks{This project was supported by funding from the Ford-Stanford Alliance and a gift from Mercedes-Benz Research \& Development North America.}
\thanks{We thank Marcos Paul Gerardo Castro, Michael Hafner, and Eric Tseng for the insightful discussions.}
\thanks{$^{1}$Department of Aeronautics and Astronautics, Stanford University (e-mail: \{hdelecki, mitkina, blange, mykel\}\!@stanford.edu).}%
\thanks{$^{2}$Department of Computer Science, Stanford University (e-mail: ransalu@stanford.edu).}%
}

\newcommand{\ph}[1]{{\textbf{#1}:}} %

\begin{document}

\maketitle
\thispagestyle{empty}
\pagestyle{empty}

\begin{abstract}



Autonomous vehicles (AVs) rely on environment perception and behavior prediction to reason about agents in their surroundings. These perception systems must be robust to adverse weather such as rain, fog, and snow. However, validation of these systems is challenging due to their complexity and dependence on observation histories. This paper presents a method for characterizing failures of LiDAR-based perception systems for AVs in adverse weather conditions. We develop a methodology based in reinforcement learning to find likely failures in object tracking and trajectory prediction due to sequences of disturbances. We apply disturbances using a physics-based data augmentation technique for simulating LiDAR point clouds in adverse weather conditions. Experiments performed across a wide range of driving scenarios from a real-world driving dataset show that our proposed approach finds high likelihood failures with smaller input disturbances compared to baselines while remaining computationally tractable. Identified failures can inform future development of robust perception systems for AVs.

\end{abstract}

\section{Introduction}\label{sec:intro}
Autonomous vehicles (AVs) rely on perception systems to reason about critical information in their surroundings, such as the presence of other vehicles and their future behavior. These systems must perform reliably in a wide variety of real-world scenarios that may not be be present during development or may occur infrequently. Reliable perception performance in adverse weather continues to be a challenge for state-of-the-art perception systems~\cite{mirzaRobustnessObjectDetectors2021, zangImpactAdverseWeather2019}. Identifying how and when these systems fail in such conditions is critical to the development and eventual deployment of autonomous systems in human environments.

Many state-of-the-art AV perception systems rely on independent modules to perform object detection, tracking, and trajectory prediction~\cite{arnoldSurvey3DObject2019, vanbrummelenAutonomousVehiclePerception2018, mozaffariDeepLearningBasedVehicle2022}. These modules may use data-driven or classical methods. Verifying the input-output properties of an individual module is not sufficient to guarantee good performance, since small errors in tasks like object detection may be magnified in downstream modules like trajectory prediction. Additionally, these systems rely on sequential observations of the environment, meaning that failures are associated with trajectories of detected objects. Searching for failures over trajectories is difficult due to the high-dimensionality of the search space.


\begin{figure}[t]
    \centering
    \includegraphics[width=1.0\columnwidth]{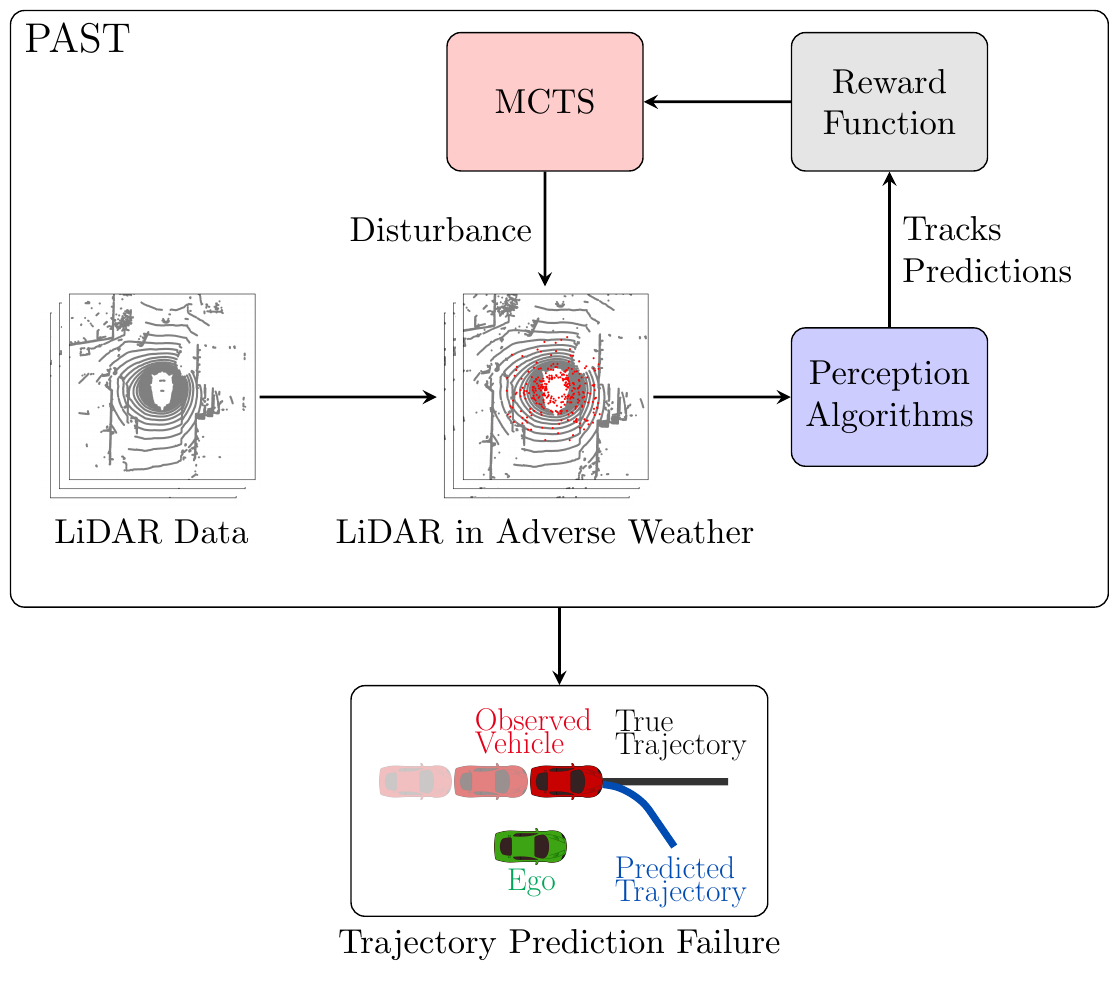}
    \caption{Our proposed \textit{PAST} framework for validation of tracking and prediction stages of an AV perception systems under adverse weather conditions. Red points indicate disturbances to the perception system caused by LiDAR beams reflecting off of airborne particles. MCTS controls how sequences of disturbances are added to the system to cause failures in perception tasks that depend on detections aggregated over time, such as object tracking and trajectory prediction. PAST seeks to find the most likely failures in these sequential perception system tasks.}
    \label{fig:ast-diagram}
\end{figure}

Many previous approaches to validation of perception systems draw on the idea of an adversarial attack on object detection~\cite{xieAdversarialExamplesSemantic2017, xiangGenerating3DAdversarial2019, eykholtPhysicalAdversarialExamples2018}. These methods solve an optimization problem to find a small perturbation or physical object that leads to an incorrect object detection when added to the base input. However, they do not consider the impact that such perturbations could have on downstream perception tasks that depend on a history of detections, such as object tracking and prediction. There have been several studies on the impact of weather disturbances on sensing and object detection~\cite{kutilaAutomotiveLiDARPerformance2018, liLidarAutonomousDriving2020}. The performance of state-of-the-art object detectors can be severely degraded in the presence of rain, fog, or snow~\cite{mirzaRobustnessObjectDetectors2021}. These studies primarily consider the robustness of sensing capabilities and object detection, but not the potential impact of weather on downstream perception tasks.

This work focuses on the validation of LiDAR-based perception systems in adverse weather. Research in validation of complex systems has led to the development of adaptive stress testing (AST), which uses reinforcement learning to find the most likely ways that a system can fail~\cite{leeAdaptiveStressTesting2015}. AST is used to validate black-box systems, where no knowledge of the system under test is available. The reinforcement learning agent adds sequences of disturbances to the system to try to cause failure. However, existing work on AST has primarily considered stress testing of decision-making systems, where the goal is to drive the true state of the system to failure. In contrast, our goal is to drive a system's estimated state to a failure condition.

Our approach frames perception system validation as a sequential decision making problem. We extend the AST framework by using reinforcement learning to search for sequences of disturbances that lead to perception system failure. Our method is able to find sequences of disturbances introduced by models of adverse weather that cause the perception system to incorrectly track and predict the behavior of surrounding vehicles. 

This paper makes the following contributions: We present a formulation of AST for perception systems which we call Perception AST (PAST), illustrated in \cref{fig:ast-diagram}. We apply PAST to find failures in a state-of-the-art AV LiDAR-based perception system under adverse weather conditions. Our experiments over many driving scenes show that our method can find more likely failures compared to other approaches.



\section{Related Work}\label{sec:relatedwork}
\subsection{Validation Methods}

Safety validation is the process of ensuring safe operation of a system in an operating environment~\cite{corsoSurveyAlgorithmsBlackBox2021}. Methods such as statistical model checking have been used to estimate the probability that a perception system's estimates comply with some specifications~\cite{barbierValidationPerceptionDecisionMaking2019}. Recently, AST has been applied to autonomous systems to search for the most likely ways that decision-making systems can fail~\cite{leeAdaptiveStressTesting2015, korenAdaptiveStressTesting2019, julianValidationImageBasedNeural2020}. This work extends the AST formulation to search for the most likely failures in perception systems rather than planning and control systems.


\subsection{Adversarial Attacks}
There has been a significant amount of work investigating robustness of image-based object detection with deep neural networks~\cite{xieAdversarialExamplesSemantic2017, yuanAdversarialExamplesAttacks2019}. These methods typically introduce local or global perturbations on inputs that cause networks to incorrectly classify or miss objects that would otherwise be detected. For more on these methods, refer to the survey by \citeauthor{akhtarThreatAdversarialAttacks2018}~\cite{akhtarThreatAdversarialAttacks2018}. Generative adversarial networks (GANs) have also been used to generate attacks on object detection and to identify out-of-distribution examples in object detection~\cite{weiTransferableAdversarialAttacks2019, nitsch2021out}. These methods require ample data to generate realistic attacks which may not be available and do not consider temporal sequences of observations.

Adversarial attack methods for image-based detectors have inspired methods for point-cloud based representations~\cite{liuExtendingAdversarialAttacks2019}. Point-cloud adversarial attacks can use similar perturbations as images, such as adding noise to point positions or completely removing some points from the input~\cite{xiangGenerating3DAdversarial2019, wickerRobustness3DDeep2019}. These adversarial methods can quantify the robustness of deep-learning based object detectors, but do not consider the impact of disturbances on downstream temporal tasks such as object tracking and prediction.



\subsection{Impact of Weather on Perception}

Prior works studied the impact of adverse weather such as rain, fog, and snow on LiDAR data~\cite{zangImpactAdverseWeather2019, liLidarAutonomousDriving2020, kutilaAutomotiveLiDARPerformance2018}. Adverse weather tends to decrease the maximum detection range, add noise to range measurements, and introduce backscattering where LiDAR beams reflect off of particles in the air. Further studies have investigated building and applying models of adverse weather conditions to evaluate the robustness of object detection~\cite{bijelicSeeingFogSeeing2020, mirzaRobustnessObjectDetectors2021, kilicLidarLightScattering2021}. We draw on these models to simulate the impact of adverse weather conditions on LiDAR data. Specifically, these weather models provide controllable disturbances to perception system input data that AST uses to induce failures.



\section{Method}\label{sec:method}

This section describes our Perception Adaptive Stress Testing (PAST) formulation. First, we overview the typical components of an AV perception system. Next, we define the objective function for PAST, and show how it can be framed as a Markov decision process (MDP).



\subsection{Problem Setting}
This work considers LiDAR-based AV perception systems with modular components for object detection, tracking, and prediction. At each time step, the perception system takes as input a 3D LiDAR point cloud and produces a list of object tracks and motion predictions. We assume that object tracks are represented by 3D bounding boxes with associated position, velocity, and orientation. Predictions are made for all tracked vehicles that are not parked. Trajectory predictions for individual objects are represented by a time series of future positions. Our goal is to find likely sequences of disturbances to the input LiDAR data that cause large errors, or failures, in the object tracks and predictions.


\subsection{Adaptive Stress Testing for Perception Systems}
AST is a configuration of model-free reinforcement learning for black-box system validation~\cite{leeAdaptiveStressTesting2015}. Rather than learning a policy that optimizes an agent's performance, AST optimizes disturbances to the environment that cause an agent to fail. Previous applications of AST have focused on validation of decision-making systems, where the goal is to drive the true state of a system to failure. We present a formulation of AST for the validation of perception systems called Perception AST (PAST). In contrast to AST, the goal of PAST is to drive the estimated state of a system to failure. 

We define a generative black box simulator $\mathscr{S}$ comprised of time-series sensor data, a stochastic disturbance model, and a perception system. We denote the hidden internal state of the system at time $t$ as $s_t$. The internal state represents the perception system's belief or state estimate about the true states and future trajectories of surrounding agents. The simulation is stepped through time by drawing a random next state, where the sampling is pseudorandomly generated from a provided seed. In our simulator, we first sample an input disturbance from the disturbance model and then use the perturbed data to update the perception system's state estimate. The simulator exposes the following four functions for simulation control:

\begin{itemize}
    \item $\Initialize(\mathscr{S})$: Resets the simulator $\mathscr{S}$ to an initial state $s_0$. This function resets the internal state of the perception system and the sensor data.
    
    \item $\Step(\mathscr{S}, a)$: Steps the simulation $\mathscr{S}$ in time by updating the perception system with a sample from the disturbance model. The randomness in the disturbance is controlled by the psuedorandom seed $a$.  This function returns the likelihood of the transition, which is the likelihood of the sampled disturbance. 
    
    \item $\IsTerminal(\mathscr{S})$: Returns true if simulator $\mathscr{S}$ has reached the end of the simulation horizon and false otherwise. 
    
    \item $\IsFailure(\mathscr{S})$: Returns true if the perception system in simulator $\mathscr{S}$ has reached a failure.
\end{itemize}

The goal of PAST is to find the most likely sequence of disturbances generated by seeds $a_{0:t}$ such that $\IsFailure(\mathscr{S})$ is true. Equivalently, we can formulate the objective as finding the sequence of pseudorandom seeds that maximizes the joint probability of disturbances subject to the constraint that the disturbances lead to a failure.


Following the AST framework, we recast this problem into an MDP. An MDP is defined by the tuple $\left(\mathcal{S}, \mathcal{A}, R, T \right)$~\cite{kochenderferAlgorithmsDecisionMaking2022}. An agent chooses an action $a \in \mathcal{A}$ based on a state $s \in \mathcal{S}$ and receives a reward $r$ based on the reward function $R(s,a)$. The state transitions stochastically to the next state $s'$ according to the transition model $T(s'~\mid~s,~a)$. 
In the PAST MDP, actions correspond to psuedorandom seeds for the generation of disturbances to the perception input data. Recall that since the actions $a$ control the stochastic transition of the system, the sequence of actions uniquely determines the system's state $s$. The transition model for the MDP is represented by the $\Step$ function exposed by the simulator, which takes in an action and returns the probability of the transition.

The reward function is designed to be equivalent to maximizing the joint probability of all actions, assuming that each action is independent. The functions $\IsFailure$ and $\IsTerminal$ exposed by the simulator are used to calculate the reward according to:
\begin{equation}
    R(s, a) = \begin{cases}
    0 & \text{if \IsFailure($s$)}\\
    \log p(a) & \text{else if not \IsTerminal($s$)}\\
    -\alpha & \text{otherwise}\\
    \end{cases}
    \label{eq:ast-reward}
\end{equation}
where $\alpha$ is a large term that penalizes cases where a terminal state is reached that is not a failure. Assuming actions are independent, the agent in the MDP is maximizing the sum over $\log p(a_t)$, which is equivalent to maximizing the product over $p(a_t)$ or the probability of the action sequence.

The goal of a PAST agent is to maximize its expected utility by finding a policy $\pi$ that specifies an action $a=\pi(s)$. The utility of following a policy $\pi$ from state $s$ is given by the value function:
\begin{align}
    V^{\pi}(s) = R(s, \pi(s)) + \gamma \sum_{s'} T(s' \mid s, \pi(s)) V^{\pi}(s')
    \label{value-function}
\end{align}
where $\gamma$ is the discount factor that controls the weight of future rewards. Algorithms such as Monte Carlo Tree Search (MCTS) can be used to find an optimal policy.
MCTS is an online sampling-based algorithm that can be used to find solutions to MDPs~\cite{browneSurveyMonteCarlo2012}. MCTS builds a search tree by sampling the state and action spaces, and estimates the value of states and actions through forward simulation. This work uses a variant of MCTS with double progressive widening (DPW)~\cite{couetoux_continuous_2011}. DPW regulates the branching factor in the state and action spaces to prevent the number of children in the search tree from exploding when the state or action spaces are very large. 

\begin{figure}[!tbp]
  \centering
  \subfloat[Tracking failure]{\includegraphics[width=0.45\linewidth]{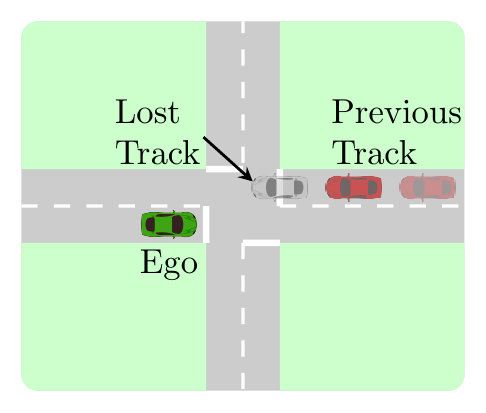}\label{fig:track-failure-illustration}}
  \hfill
  \subfloat[Prediction Failure]{\includegraphics[width=0.45\linewidth]{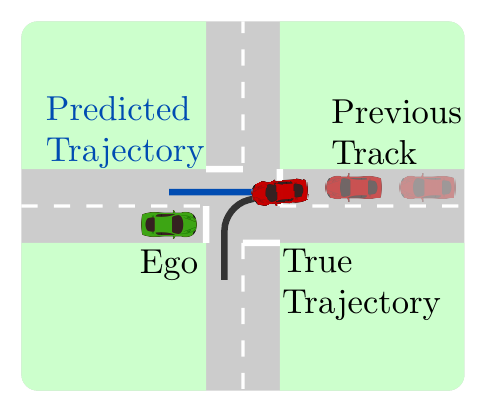}\label{fig:pred-failure-illustration}}
  \caption{Illustration of tracking failure (left) and prediction failure (right). In both images, the ego vehicle is in blue and the observed vehicle is in red.}
  \label{fig:failure-defs}
\end{figure}

\section{Experiments}\label{sec:experiments}

\begin{figure*}[t!]
    \centering
    \includegraphics[width=2\columnwidth]{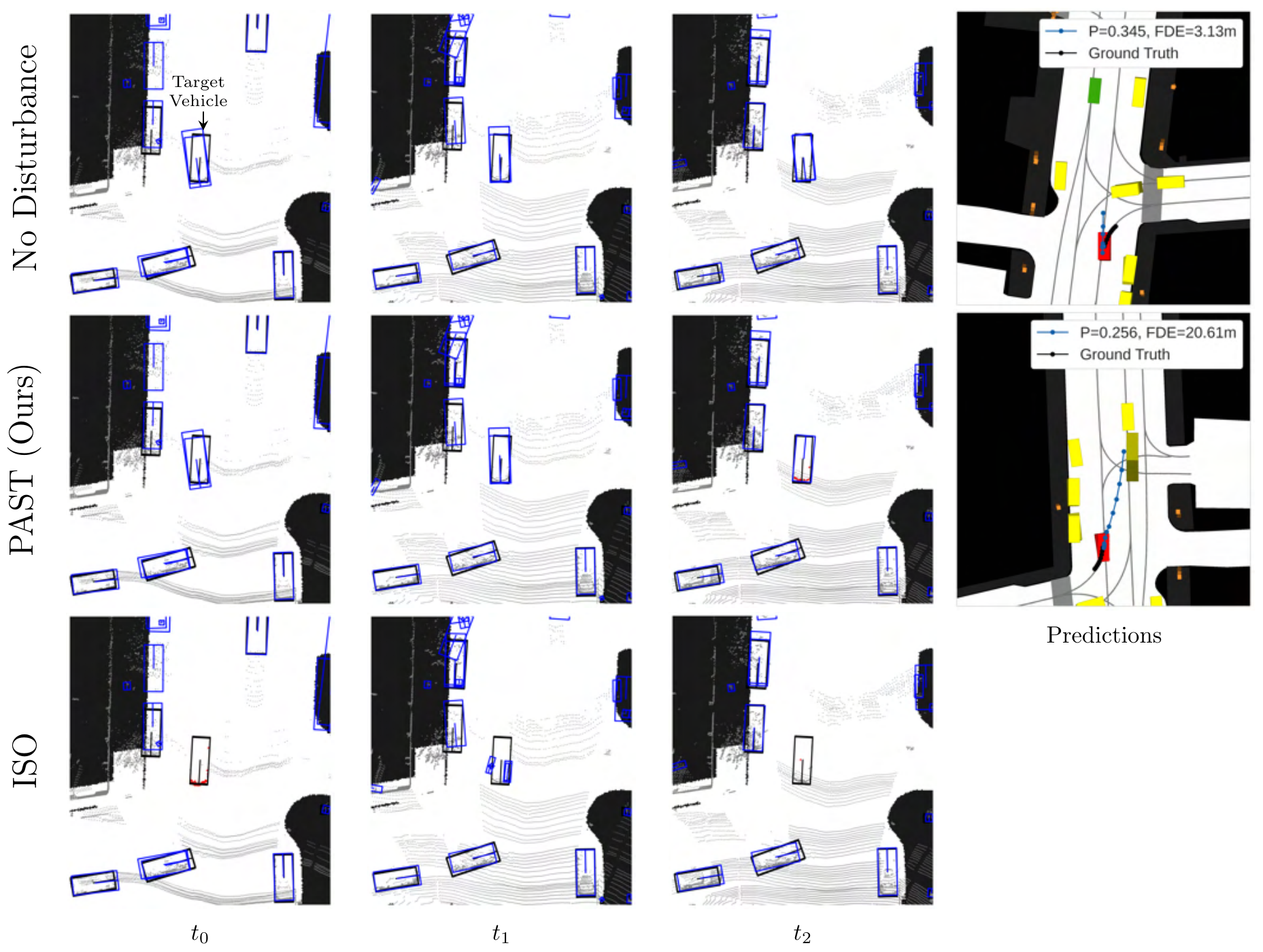}
    \caption{Detections (blue) and ground truth (black) bounding boxes leading up to the failure discovered in the single object perception PAST experiment. The target vehicle is in the center of each detection frame. The solid line inside the box indicates heading. PAST does not remove any points until $t_3$, where it removes $15$ points from the front of the target vehicle. This disturbance causes a large error (almost \SI{180}{\degree}) in the detected heading of the target vehicle. Under this simple disturbance identified by PAST, the target vehicle is predicted to be travelling in the opposite direction as the ground truth. The ISO baseline removes points at each time step that prevent the target from being detected. Since there are no detections of the target vehicle with ISO, the perception system can't make a trajectory prediction. The target vehicle is shown in red in the predictions. AST is able to find a more likely path to failure by considering disturbance trajectories.}
    \label{fig:simple-failure}
\end{figure*}

In this section we present the experimental setup used to illustrate the performance of our proposed method. We describe specific data sources, perception system components, and failure definitions for each experiment. We consider two experimental settings. We first illustrate the effectiveness of our approach at finding failures in the perception of a single vehicle. Then we consider an at scale experiment using an adverse weather disturbance model.

\subsection{Experimental Setup}
We perform validation of an AV perception system using data from the real-world driving dataset, nuScenes~\cite{caesarNuScenesMultimodalDataset2020}. 
The dataset contains many hours of real-world driving data divided into \SI{20}{\second} long scenes. Each scene contains 32-beam LiDAR sweeps at \SI{20}{\hertz} and ground truth 3D object bounding boxes. We only consider the $123$ scenes from the validation split that were recorded in clear weather conditions. We use the validation split to avoid any issues that may be caused by overfitting in perception modules. Our algorithm does not involve training or tuning hyperparameters based on this data. We treat each of these scenes as an independent validation case, in which we stress test the system to find a likely sequence of disturbances that lead to failure. Each scene is simulated by stepping through time, perturbing the recorded point cloud, and updating the perception system.

\subsection{Perception System Under Test}
We consider validation of state-of-the-art detection, tracking, and prediction modules that are commonly used to baseline nuScenes perception tasks. Object detection is performed by the PointPillars architecture~\cite{langPointPillarsFastEncoders2019,openpcdetdevelopmentteamOpenPCDetOpensourceToolbox2020}. Multi-object tracking is performed by AB3DMOT, which uses a Kalman filter to maintain tracks of 3D bounding box observations~\cite{wengAB3DMOTBaseline3D2020}. Finally, trajectory prediction is performed by CoverNet~\cite{phan-minhCoverNetMultimodalBehavior2020}, which makes predictions based on a pre-determined set of candidate trajectories.


\subsection{Failure in Single Object Perception}
As an illustrative example, we consider the case where disturbances only interfere with the perception of a particular vehicle in the scene, which we refer to as the ``target vehicle''. We hand select a target vehicle from the dataset, and define a simplified disturbance model that only impacts points inside the target vehicle's bounding box.

\ph{Disturbance Model} We assume that every LiDAR point in the bounding box is associated with an independent Bernoulli random variable representing whether the point will be removed. All points have the same probability of being removed, $\theta$. Given a random sample generated with seed $a$ for all $m$ available points, the likelihood of the disturbance is:
\begin{equation}
    p(a) = \theta^n (1-\theta)^{m-n}
    \label{eq:toy-likelihood}
\end{equation}
where $n$ is the number of points selected to be removed. For demonstration purposes, we use $\theta=0.1$.

\begin{table*}
\begin{center}
\caption{PAST and baseline results on the nuScenes dataset. Mean and standard error values are provided for the local disturbance magnitude $\delta$, global disturbance magnitude $\Delta$, and trajectory length. Higher failure rate indicates better performance, while lower values of disturbance magnitudes and trajectory length are desired. PAST is able to find many likely failures in both tracking and prediction.}
\label{table:ast-results}
    \begin{tabular}{@{}llrrrrr@{}}
    \toprule
    Failure Criteria & Method & Failure Rate (\%) & $\delta$ (\%) & $\Delta$ (\%) & Trajectory Length (-)\\ 
    \midrule
    Tracking and Prediction & ISO (Baseline) & 69.1 &  8.1 $\pm$ 2.1  & 0.11 $\pm$ 0.20 & 5.75 $\pm$ 0.57\\
    & MC (Large) & 25.1  & \num{4.68e-2} $\pm$ \num{7.6e-3} & 3.94 $\pm$ 0.35 & 12.2 $\pm$ 1.2 \\
    & MC (Small) & 22.1 & \num{3.08e-2} $\pm$ \num{4.5e-3} & 3.54 $\pm$ 0.54 & 13.9 $\pm$ 1.3\\ 
    & PAST (Large) & 51.2  & \num{4.06e-2} $\pm$ \num{1.2e-2} & 4.10 $\pm$ 0.42 & 7.54 $\pm$ 0.62 \\
     & PAST (Small) & 31.7 & \num{2.71e-2} $\pm$ \num{8.2e-3} & 4.00 $\pm$ 0.53 & 8.63 $\pm$ 0.57\\ 
     \midrule
    Prediction & MC (Prediction) & 21.9 & \num{5.18e-2} $\pm$ \num{7.1e-3} & 3.19 $\pm$ 0.57 & 16.5 $\pm$ 1.1 \\ 
    & PAST (Prediction) & 29.3 & \num{7.70e-2} $\pm$ \num{1.5e-2} & 3.41 $\pm$ 0.55 & 9.46 $\pm$ 0.71 \\ 
    \bottomrule
    \end{tabular}
    \label{1}
\end{center}
\end{table*}

\ph{Failure Definition} We define a failure in perception for this simple example in terms of the target vehicle's estimated track and trajectory prediction. These definitions are illustrated in \cref{fig:failure-defs}. We define failures in tracking to occur when the error in the estimated position exceeds~\SI{2}{\meter}, or when the track associated with the target vehicle is lost. We define a failure in prediction to occur when the final displacement error exceeds \SI{15}{\meter} in CoverNet's most likely predicted trajectory. We chose this threshold to capture cases where the predicted intent of a vehicle is significantly different from ground truth. In practice, these failure thresholds should be decided by vehicle manufacturers and policy makers.



\subsection{At Scale Validation in Adverse Weather}
In the following set of experiments, we perform validation on all 123 scenes from the nuScenes dataset with disturbances based on an adverse weather model. These experiments demonstrate the ability of our method to scale validation of a state-of-the-art perception system over a wide range of driving scenes.

\ph{Disturbance Model} We use the LiDAR Light Scattering Augmentation (LISA) software package to model the effects of adverse weather on LiDAR data~\cite{kilicLidarLightScattering2021}. LISA provides methods to augment LiDAR measurements with physics-based models of rain, fog, and snow. In our experiments, we focus on disturbances due to rain. LISA takes as input a rain rate and returns a new point cloud with simulated rain effects. The algorithm generates a new point cloud by randomly sampling from physics-based distributions to remove points, add range noise, and reflect points. For use within our PAST framework, we modify LISA to accept a random seed and to compute the log-likelihood of sampled disturbances.


\ph{Failure Definition} The definitions of perception failure are slightly different in adverse weather experiments to more accurately reflect how a real-world perception system might be used. Here, failures in prediction occur when the minimum final displacement error exceeds \SI{15}{\meter} over CoverNet's top five most confident predictions. The definition of tracking failures is the same as in the single object perception experiment. We only check for failures that emerge due to the adverse weather disturbances. In particular, if a failure criterion is met for a vehicle without disturbances, we do not terminate PAST for this failure event.

We perform PAST over all $123$ scenes using three different configurations. The first configuration we call PAST (Large). PAST selects from three relatively heavy rain rates of \num{20}, \num{30}, and \SI[per-mode = symbol]{40}{\milli \meter \per \hour}. We consider more mild disturbances in PAST (Small), where we consider rain rates of \num{5}, \num{10}, and \SI[per-mode = symbol]{15}{\milli \meter \per \hour}. The last experiment PAST (Prediction) uses the same smaller rain rates, but only considers failures in prediction. PAST (Prediction) demonstrates the flexibility of PAST to find failures under different criteria. For all PAST experiments, we use MCTS with a maximum of $2,000$ iterations.


\subsection{Baselines}
As a baseline for our approach, we consider a modified version of the Iterative Salience Occlusion (ISO) algorithm, an adversarial attack on 3D object detection~\cite{wickerRobustness3DDeep2019}.  ISO uses latent feature spaces of 3D object detectors to identify sets of critical points, or points in the input space that contribute the most to the network's identification of an object. ISO iteratively removes these critical points from the input until the network misclassifies the result. The original ISO algorithm was created for single object detection. We use a modified version of the ISO algorithm for the multi-object detection task in AV perception.During simulation, we run ISO each time step until a maximum number of iterations is reached or a vehicle is misclassified. Rather than checking for misclassification of a single object, we check for misclassification of any agent of a specific class, such as `car'. After termination of ISO, we use the new point cloud that it returns to update the perception system's state estimate. Adversarial attack methods like ISO struggle to stay computationally tractable when considering multiple time steps and very large point clouds. We restrict ISO to consider LiDAR points inside the ground truth bounding boxes of other vehicles and set a limit of $100$ iterations to ensure computational tractability in our experiments.

We also baseline our approach using Monte Carlo (MC) random search with the adverse weather disturbance model to confirm that PAST is able to maximize its objective. Random search selects actions at random using the same number of iterations as MCTS and returns the maximum likelihood failure discovered. This baseline is repeated for each of the large rain rate, small rain rate, and prediction-only experiment cases that we consider for PAST.

\section{Results}
Experiments were conducted on a desktop with 32GB of RAM, an Intel i7-7700K CPU, and an NVIDIA GeForce GTX 1080 Ti GPU. PAST was implemented using AST-Toolbox, an open-source software package for designing and running AST experiments.\footnote{https://github.com/sisl/AdaptiveStressTestingToolbox} We first consider an experiment involving perception of a single object in a scene. Next, we perform experiments over many driving scenes using an adverse weather disturbance model, demonstrating our method's ability to scale. Our code is available at: \href{https://github.com/sisl/PerceptionAdaptiveStressTesting}{https://github.com/sisl/PerceptionAdaptiveStressTesting}.

\ph{Failure in Single Object Perception}
The object detections and predictions for the single object perception experiment are illustrated in \cref{fig:simple-failure}. In the detections, the target vehicle is shown in the center of the frame. The target vehicle is making a left turn while becoming occluded with respect to the ego vehicle. In the predictions, the target vehicle is shown in red. The perception system successfully outputs good tracks and predictions without disturbances as seen in the first row.

The most likely failure event according to PAST is shown in the second row in \cref{fig:simple-failure}. The most likely failure in this experiment corresponds to the failure that results from the fewest total number of points removed.
At the beginning of its search, PAST removes points at each time step. Over successive iterations, PAST seeks to maximize the expected reward, or minimize the number of points removed. PAST discovers that a failure can occur by only removing $15$ points at the third time step (as shown in \cref{fig:simple-failure}). The ISO baseline removes $112$ LiDAR points in total, removing points at every time step, resulting in the target vehicle never being detected by definition of the approach. By reasoning over trajectories of disturbances, PAST is able to find a failure trajectory that removes significantly fewer points than the ISO baseline.

\begin{figure}[tbp!]
    \centering
    \includegraphics[width=0.9\columnwidth]{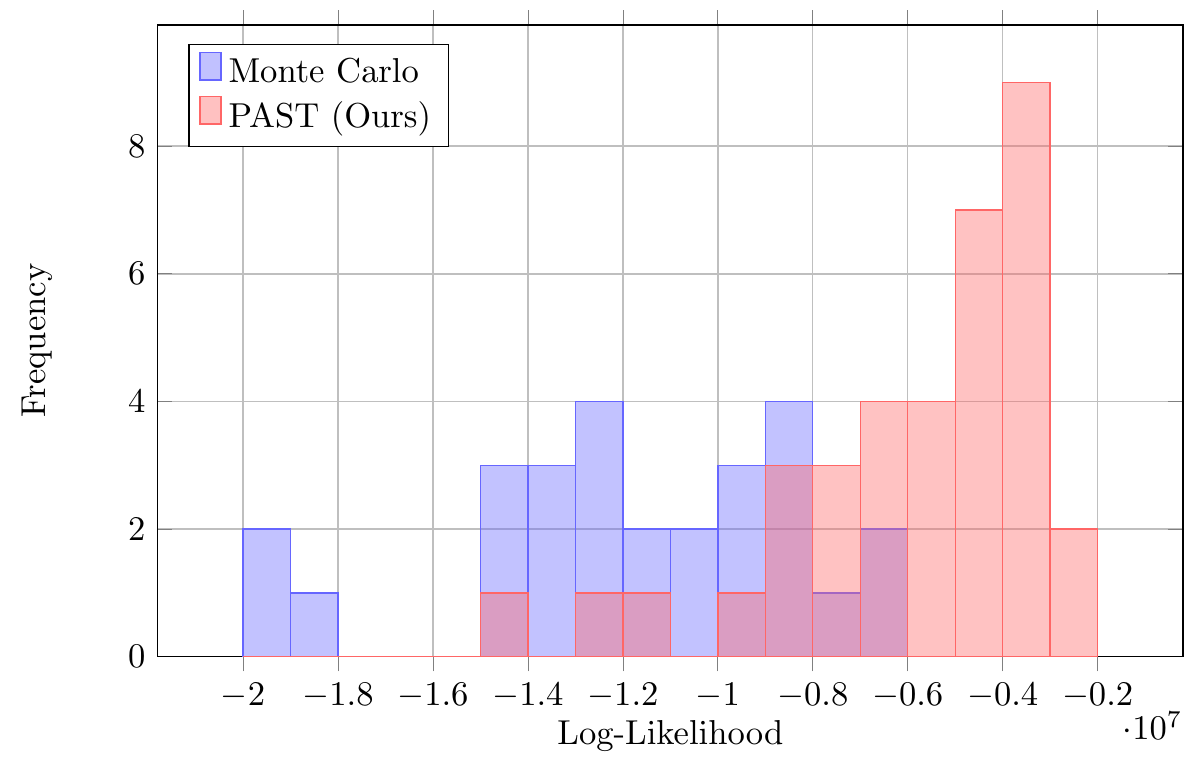}
    \caption{Histogram of the total log-likelihood of failures in trajectory prediction found using Monte Carlo random search and PAST. PAST successfully finds more likely failures than random search.}
    \label{fig:mc-past-prediction-hist}
\end{figure}


\ph{At Scale Validation in Adverse Weather}
For the at scale experiment, we compare the performance of our method against baselines in terms of failure rate, mean local disturbance magnitude $\delta$, global disturbance magnitude $\Delta$, and mean trajectory length. Failure rate refers to the proportion of scenes that the method was able to find failures in. Local disturbance magnitude $\delta$ is the proportion of points removed from inside the bounding box of the vehicle involved in the failure. Global disturbance magnitude $\Delta$ is the proportion of points moved or removed in the whole point cloud. Trajectory length refers to the number of observations of the failure vehicle leading up to the failure event.

A summary of experimental results for PAST and baseline methods over the nuScenes validation split is shown in \cref{table:ast-results}. The first five rows in \cref{table:ast-results} show results for experiments considering failures in tracking and prediction. PAST outperforms MC in failure rate as well as in trajectory length.
The baseline ISO algorithm is substantially more aggressive in the number of points removed locally than our algorithm, resulting in a higher failure failure rate when considering failures in tracking and prediction.
For ISO to be tractable at the scale of these experiments, it was restricted to consider points inside ground truth bounding boxes. ISO tends to remove a higher proportion of LiDAR points associated with specific vehicles to introduce poor detections. In contrast, while PAST adds a disturbance over the whole point cloud, this disturbance is relatively small with respect to the ground truth vehicles as demonstrated by the $\delta$ metric. Therefore, PAST can scale validation to larger datasets and add smaller disturbances by considering longer failure trajectories.

When considering failures in tracking and prediction, all of the failures found by PAST and the baseline methods occur in object tracking. To find failures in prediction, there must first be tracks to predict. Since the disturbances cause poor detections, it is more difficult to maintain good object tracks and the perception system fails in tracking before failures in prediction can manifest. The last two rows of \cref{table:ast-results} show results considering only failures in prediction. The ISO algorithm was unable to find failures in prediction. Both MC (Prediction) and PAST (Prediction) were successful, since they are able to consider longer failure trajectories. PAST finds more failures in predictions than MC because it is able learn sequences of actions that maximize the PAST objective function. \cref{fig:mc-past-prediction-hist} shows a histogram comparing the likelihood of failures found in MC (Prediction) and PAST (Prediction). In addition to being able to find more failures than MC, PAST also finds failures that are shorter and more likely. The high performance of PAST compared to random search suggests that PAST is maximizing its objective function.
\begin{figure*}[t!]
    \centering
    \includegraphics[width=2\columnwidth]{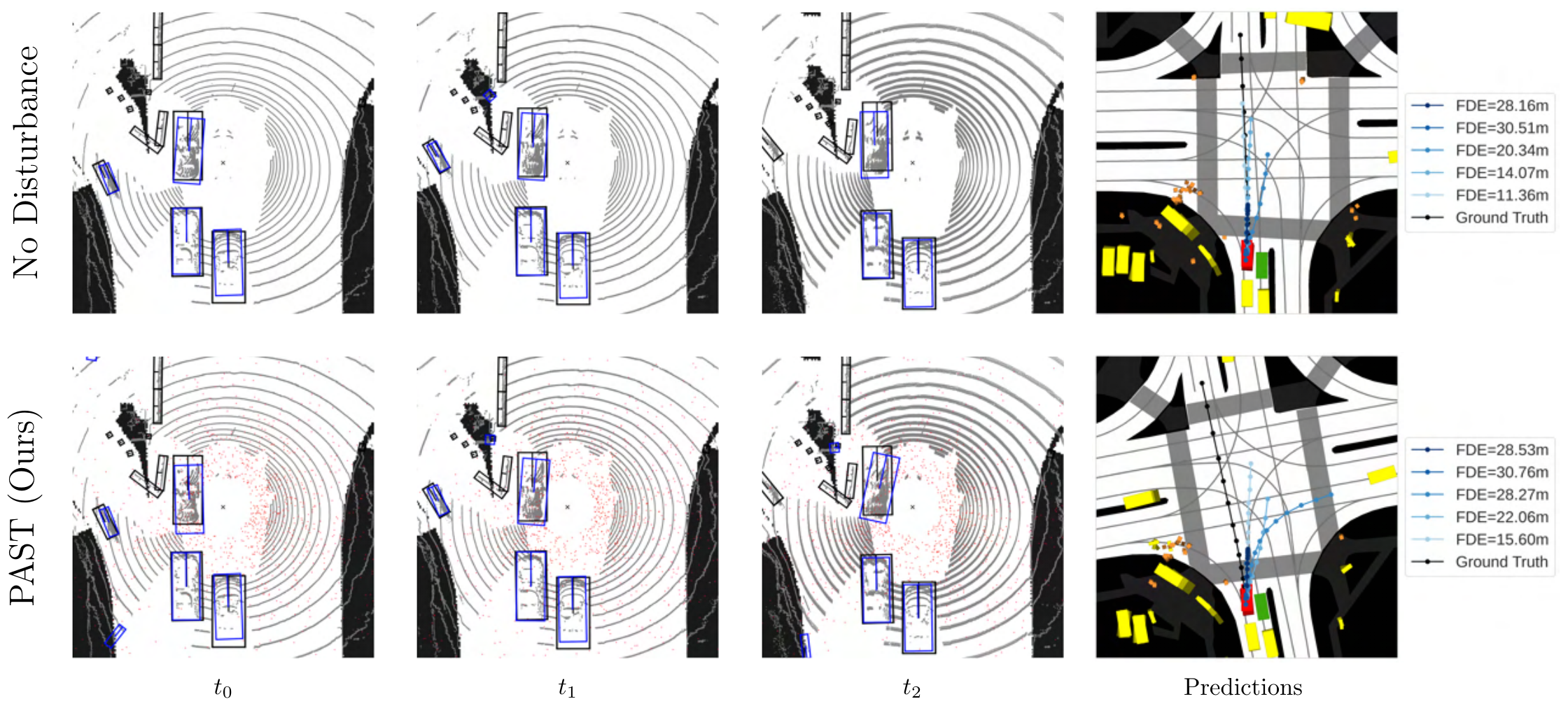}
    \caption{Comparison of detections and predictions without disturbances (top) and with disturbances found by PAST (bottom). Red points indicate points reflected due to the adverse weather model. Detected bounding boxes are shown in blue and the ground truth is shown in black. With the adverse weather disturbance, the detections of the vehicle just to the left of the ego vehicle appear to be turning right. Predictions are shown in the last column with the ego vehicle in green and the predicted vehicle in red. The observations under disturbances cause the predicted trajectory to be turning right and in front of the ego vehicle when the vehicle is actually accelerating straight ahead. This type of failure could cause the AV to brake unnecessarily in traffic, causing a potentially unsafe driving scenario. Results for ISO are not shown since it is unable to find a failure.}
    \label{fig:nusc-prediction-failure}
\end{figure*}

An example failure trajectory is illustrated in \cref{fig:nusc-prediction-failure}. In this scene, the ego vehicle is stopped at a light with a few surrounding vehicles. The vehicle just to the left of the ego begins to accelerate forward. Without disturbances, this behavior is predicted accurately. However, the rain disturbance identified by our PAST method makes it appear to the perception system that the vehicle is turning right and into the ego vehicle's lane, cutting the ego vehicle off. This incorrect prediction about the other vehicle's future trajectory could lead to undesirable performance in the AV, such as an unnecessary hard braking maneuver.

PAST required \SI{90}{\minute} for a single scene on average, while Monte Carlo and ISO averaged \SI{60}{\minute} and \SI{200}{\minute}, respectively. While ISO finds shorter paths to failure than MC or PAST, its aggressive local disturbances are not likely to occur in the real world. Considering smaller disturbances with ISO becomes computationally intractable due to the high dimensionality of the search space. PAST is able to find failures in prediction and tracking that are relatively small while remaining computationally tractable.

\section{Conclusion}
\label{sec:conclusion}
AVs rely on perception and prediction to reason about their surroundings. Identifying how and when these systems fail in adverse weather conditions is critical to the development and deployment of autonomous systems in human environments. This work presents PAST, a method for validation of LiDAR-based perception systems that uses reinforcement learning to find likely failures. The method was applied to the validation of a perception system in adverse weather using real-world LiDAR data. The results showed that the proposed PAST method tractably finds likely disturbances that introduce large errors into the tracking and prediction of other vehicles across a range of driving scenarios. A key future research direction is to apply PAST to validate end-to-end perception and prediction methods~\cite{itkina2019dynamic,toyungyernsub2021double,lange2020attention, pnpnet2020}, enabling quantitative comparison of the robustness between modular and end-to-end perception systems under adverse weather conditions. Additionally, understanding failures in different systems could inform how to combine perception systems to balance each component's strengths and weaknesses, resulting in a more robust solution. 


\renewcommand*{\bibfont}{\small}
\printbibliography

\end{document}